\documentclass[conference]{IEEEtran}
\usepackage{times}

\usepackage[numbers]{natbib}
\usepackage{multicol}
\usepackage[bookmarks=true]{hyperref}

\usepackage{amssymb}
\usepackage{multirow}
\usepackage{makecell}
\usepackage{graphicx}
\usepackage{caption}
\usepackage{subcaption}
\usepackage{amsfonts, amsmath, amsthm}
\usepackage[table]{xcolor}
\usepackage{algorithmic}
\usepackage{algorithm}
\usepackage{marvosym}

\begin{document}

\title{ConRFT: A Reinforced Fine-tuning Method for VLA Models via Consistency Policy}

\renewcommand{\authorrefmark}[1]{\textsuperscript{#1}}

\author{\authorblockN{Yuhui Chen\authorrefmark{1}\authorrefmark{2},
Shuai Tian\authorrefmark{1}\authorrefmark{2},
Shugao Liu\authorrefmark{1}\authorrefmark{2}, 
Yingting Zhou\authorrefmark{1}\authorrefmark{2}, 
Haoran Li\authorrefmark{1}\authorrefmark{2}\textsuperscript{\Letter}, and
Dongbin Zhao\authorrefmark{1}\authorrefmark{2}\textsuperscript{\Letter}}
\authorblockA{\authorrefmark{1}SKL-MAIS, Institute of Automation, Chinese Academy of Sciences, Beijing, China}
\authorblockA{\authorrefmark{2}School of Artificial Intelligence, University of Chinese Academy of Sciences, Beijing, China\\
Email: \{chenyuhui2022, tianshuai2023, liushugao2023, zhouyingting2025, lihaoran2015, dongbin.zhao\}@ia.ac.cn}}

\maketitle

\begin{abstract}

Vision-Language-Action (VLA) models have shown substantial potential in real-world robotic manipulation. However, fine-tuning these models through supervised learning struggles to achieve robust performance due to limited, inconsistent demonstrations, especially in contact-rich environments. In this paper, we propose a reinforced fine-tuning approach for VLA models, named ConRFT, which consists of offline and online fine-tuning with a unified consistency-based training objective, to address these challenges. In the offline stage, our method integrates behavior cloning and Q-learning to effectively extract policy from a small set of demonstrations and stabilize value estimating. In the online stage, the VLA model is further fine-tuned via consistency policy, with human interventions to ensure safe exploration and high sample efficiency. We evaluate our approach on eight diverse real-world manipulation tasks. It achieves an average success rate of 96.3\% within 45–90 minutes of online fine-tuning, outperforming prior supervised methods with a 144\% improvement in success rate and 1.9x shorter episode length. This work highlights the potential of integrating reinforcement learning to enhance the performance of VLA models for real-world robotic applications. Videos and code are available at our project website \href{https://cccedric.github.io/conrft/}{https://cccedric.github.io/conrft/}.

\end{abstract}

\IEEEpeerreviewmaketitle

\section{Introduction}

Recent advancements in training generalist robotic policies using Vision-Language-Action (VLA) models have demonstrated remarkable capabilities in understanding and executing various manipulation tasks. These successes are primarily attributed to large-scale imitation-style pre-training and grounding with robot actions \cite{o2023open, brohan2023rt, black2024pi_0}. While pre-trained policies capture powerful representations, they often fall short when handling the complexities of real-world scenarios \cite{jones2025beyond}. Fine-tuning with domain-specific data is essential to optimize model performance for downstream tasks \cite{black2024pi_0, wang2024scaling}. While Supervised Fine-Tuning (SFT) of the VLA model using human teleoperation data remains the predominant adaptation approach, this process faces significant challenges: the model's performance heavily relies on the quality and quantity of task-specific data. However, these human-collected datasets may not consistently provide optimal trajectories due to inherent issues such as sub-optimal data and inconsistent action \cite{xu2024rldg}.

Significant progress in Large-Language-Models (LLMs) and Vision-Language-Models (VLMs) have highlighted the value of reinforcement learning as a powerful tool for bridging the gap between policy capabilities and human preference \cite{christiano2017deep, ouyang2022training, luong2024reft} or improving model reasoning \cite{pang2024iterative}. In addition, deploying reinforcement learning (RL) with task-specific reward functions to learn from online interaction data is also a promising direction \cite{ramamurthy2022reinforcement, bai2024digirl, carta2023grounding}. However, extending these insights to VLA models presents unique challenges because, unlike LLMs, VLA models necessitate direct physical interaction in real-world robotic tasks. The safety and cost constraints of collecting data in contact-rich environments demand high sample efficiency and risk-aware exploration, making a straightforward implementation of RL infeasible. Recent work has attempted to leverage RL to address the challenges faced in SFT \cite{xu2024rldg, mark2024policy}, while these methods primarily focus on utilizing RL for data augmentation or quality improvement rather than directly optimizing VLA models through RL objectives. This limits the policy’s ability to explore states out of the demonstration dataset, thus undermining the potential benefits of RL-based fine-tuning in real-world settings. 

To leverage the benefits of RL-based techniques for efficiently fine-tuning VLA models with online interaction data, we propose a reinforced fine-tuning (RFT) approach consisting of offline and online stages with a unified consistency-based training objective. While this design is similar to offline-to-online methods \cite{lee2022offline, nakamoto2024cal, zhou2024efficient}, we found that expert demonstrations' scarcity constrains their offline training performance. Motivated by insights from CPQL \cite{chen2023boosting}, we propose a unified training objective that integrates supervised learning with Q-learning in the offline stage and further fine-tunes the VLA model via consistency policy through online RL. During offline training, our approach leverages prior demonstrations and handles out-of-distribution (OOD) states, effectively extracting the policy and value function before interacting with real-world environments. In the subsequent online stage, we solve two challenges of sample efficiency and real-world safety requirements by exploiting task-specific rewards with CPQL \cite{chen2023boosting} under human interventions through Human-in-the-Loop (HIL) learning \cite{kelly2019interactive, luo2024precise}. 

Our contributions are summarized as follows: 
\begin{enumerate}
    \item We present a \textbf{Con}sistency-based \textbf{R}einforced \textbf{F}ine-\textbf{T}uning (\textbf{ConRFT}) approach, a novel pipeline with the unified training objective both for offline and online fine-tuning.
    \item By integrating offline RL with a consistency-based behavior cloning (BC) loss, we propose Cal-ConRFT, which focuses on extracting an efficient policy and value function to provide a stable initialization with a small set of demonstrations.
    \item During online fine-tuning, we propose HIL-ConRFT, which retains the same loss structure from the offline stage for rapid policy adaption while leveraging human interventions to ensure safe exploration and high sample efficiency in real-world environments.
\end{enumerate}

We evaluate our approach on eight real-world manipulation tasks, demonstrating its ability to outperform state-of-the-art (SOTA) methods. Our framework achieves an average success rate of 96.3\% after 45–90 minutes of online fine-tuning, showcasing high sample efficiency. Additionally, it outperforms SFT methods that are trained on either human data or RL policy data, with an average success rate improvement of 144\% and an episode length of 1.9x shorter. 

\section{Related Work}

\subsection{Reinforced Fine-tuning for Large Models}

RL has been widely adopted for fine-tuning LLMs and VLMs. Early works have primarily focused on RL incorporating human feedback \cite{christiano2017deep, ouyang2022training, luong2024reft, casper2023open, zhai2024fine} by learning from human preferences or by integrating task-specific rewards without explicit human preference \cite{ramamurthy2022reinforcement, bai2024digirl, carta2023grounding, kimin2021pebbla}. While many of these approaches employ on-policy algorithms (e.g., PPO \cite{schulman2017proximal}) to fine-tune pre-trained policies \cite{bai2024digirl, gupta2019relay, shao2024deepseekmath}, they typically demand large amounts of interaction data to achieve desirable performance \cite{ball2023efficient, li2024selu}. While RL has demonstrated success in many domains, it typically learns within self-generated synthetic environments rather than real-world environments. This gap prevents direct transfer for VLA models, which require real-world interaction. Our work addresses this discrepancy by developing RL frameworks tailored for efficient real-world VLA fine-tuning.

\subsection{Real-world RL Systems}

Real-world robotic RL systems require algorithms that are both sample-efficient in handling high-dimensional inputs and flexible enough to accommodate practical considerations like reward specification and environment resets \cite{luo2024precise}. Several previous methods have effectively demonstrated policy learning directly in physical environments \cite{riedmiller2009reinforcement, johannink2019residual, luo2024serl, luo2024precise}, using both off-policy \cite{tony2022offline, luo2023rlif, hu2023reboot, russell2024continuously}, on-policy \cite{zhu2019dexterous, zhuang2023robot} methods, or posing "RL as supervised learning" \cite{mark2024policy, jan2007reinforcement}. Despite this progress, many real-world RL systems still demand prolonged training sessions or require large amounts of interaction data \cite{henery2020ingredients}, which can be impractical and risk-prone in contact-rich tasks. In contrast to previous methods that train from scratch, our work focuses on utilizing pre-trained VLA models to provide high-quality policy initialization. This approach effectively mitigates unnecessary exploratory behaviors in early RL phases, thereby optimizing both policy learning efficiency and operational safety in the training process.

\subsection{Offline-to-online Methods}

Offline-to-online RL aims to leverage offline datasets to initialize a policy, which is then fine-tuned via online interactions for improved sample efficiency \cite{lee2022offline}. Existing works commonly adopt an offline pre-training stage followed by an online fine-tuning stage \cite{lee2022offline, agarwal2022reincarnating, rafailov2023moto, nakamoto2024cal}, mixing offline and online data as training proceeds. This offline-to-online pipeline is similar to our proposed two-stage fine-tuning approach that exploits pre-collected data to bootstrap policy training and then fine-tunes the policy in the real-world tasks \cite{tony2022offline}. Most offline-to-online methods assume the availability of large-scale, diverse datasets with sufficient state coverage \cite{rajeswaran2017learning, nair2018overcoming}, a condition rarely met in real-world deployments. We explore leveraging pre-trained VLA models as the base policy to enable sample-efficient policy refinement, achieving superior fine-tuning performance even under stringent demonstration data constraints.

\section{Problem Setup and Preliminaries}

We focus on fine-tuning a pre-trained VLA model for downstream tasks. Specifically, we assume access to a pre-trained VLA model $\pi_{\phi_{\mathrm{pre}}}$, which encodes high-level representations from both visual inputs (e.g., RGB images) and language instructions. In supervised fine-tuning (SFT), we aim to adapt $\phi_{\mathrm{pre}}$ to $\phi$ on the target task using a small set of labeled demonstrations while preserving the model’s general feature-extraction capability. Formally, let $\tau = (s_0, a_0, \dots, s_H)$ be a trajectory for the target task, then the VLA model fine-tuning aims to solve $\min_{\phi}\mathcal{L}(\tau, \phi)$ where $\mathcal{L}$ may be a negative log-likelihood (NLL) or a mean-squared error (MSE) measuring the discrepancy between the predicted actions and those in the demonstration. This procedure allows us to effectively leverage compressed knowledge in robotic tasks while steering the VLA model to the downstream environment.

Since demonstrations are often limited, inconsistent, and sub-optimal, preventing the policy from covering diverse states, SFT struggles in real-world, contact-rich robotic tasks. To address these issues, we formulate each robotic task as a Markov Decision Process (MDP), where the goal of RL is to find the optimal policy in the MDP, $\mathcal{M} = (\mathcal{S}, \mathcal{A}, \mathcal{P}, r, \rho, \gamma)$, where $s \in \mathcal{S}$ denotes the state space and $a \in \mathcal{A}$ denotes the action space. $\mathcal{P}(s'|s, a)$ is the environmental transition probabilities that depend on the system dynamics, and $\rho(s)$ denotes the initial state distribution. $r(s, a)$ and $\gamma \in (0, 1)$ are the reward function and the reward discount factor. The policy $\pi$ is estimated by maximizing the cumulative expected value of the reward, denoted as $V^{\pi}(s)=\mathbb{E}_{\pi}[\sum_{t=0}^{H}\gamma^tr(s_t,a_t)|s_0=s,a_t\sim\pi(s_t),s_{t+1}\sim p(\cdot|s_t,a_t)]$. The Q-function of a given policy $\pi$ is denoted as $Q^{\pi}(s,a)=\mathbb{E}_{\pi}[\sum_{t=0}^{H}\gamma^tr(s_t,a_t)|s_0=s,a_0=a,s_{t+1}\sim p(\cdot|s_t,a_t)]$. $H$ represents the maximum episode step of a trajectory. By coupling the VLA policy with the learned Q-function, RFT allows the VLA model to refine its behavior based on trial-and-error interactions and task-specific feedback.

\begin{figure*}[ht]
\centering
\includegraphics[width=\linewidth]{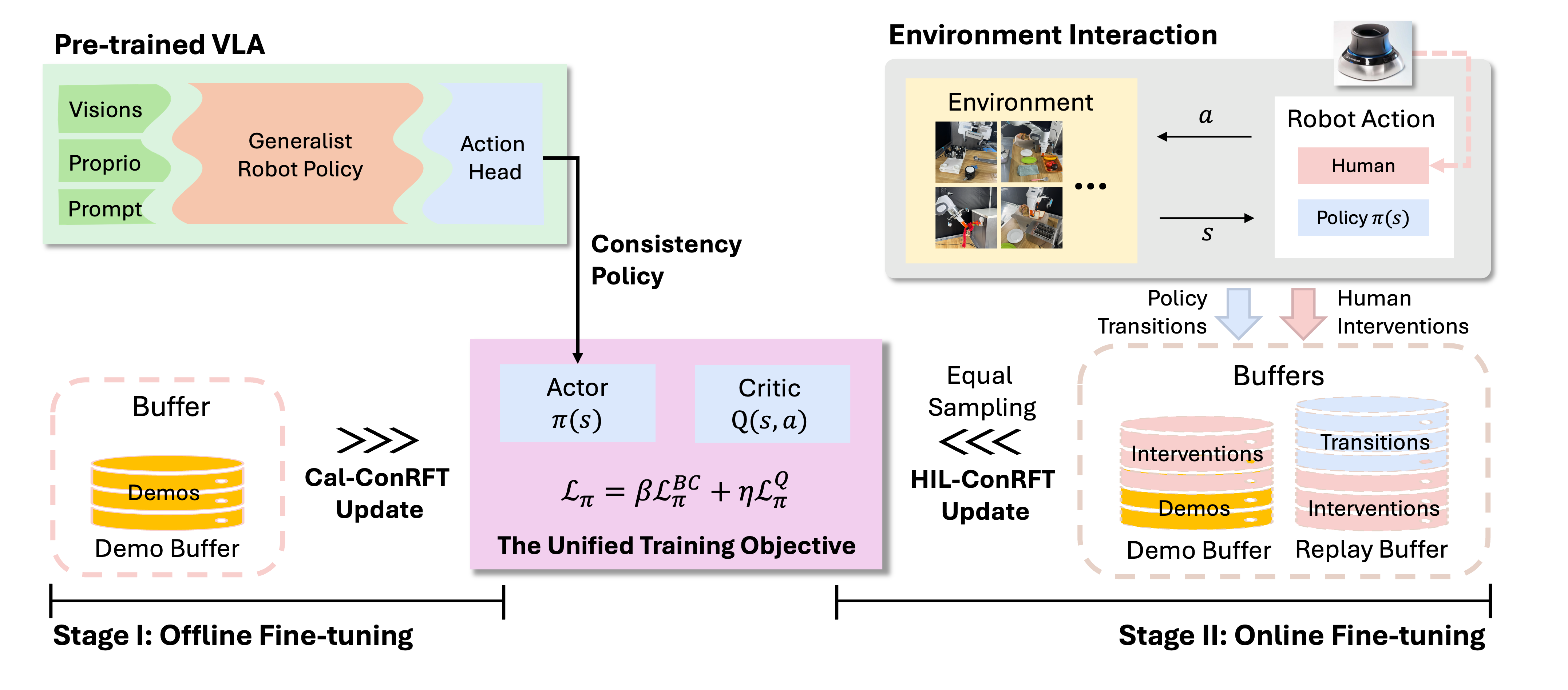}
\caption{\textbf{Overview of ConRFT.} This figure illustrates the architecture of our reinforced fine-tuning approach for a pre-trained VLA model, which comprises two stages: the offline Cal-ConRFT and the online HIL-ConRFT. Both stages use a unified consistency-based training objective. During the offline stage, we only use pre-collected demonstrations for fine-tuning. During the online stage, a human operator can intervene in the robot policy via teleoperation tools(e.g. a SpaceMouse). And we use both pre-collected demonstrations, policy transitions, and human interventions for fine-tuning. }
\label{fig:method}
\end{figure*}

\section{Method}

The proposed pipline ConRFT consists of two stages: offline fine-tuning followed by online fine-tuning to optimize robotic policies, as shown in Fig. \ref{fig:method}. In the following sections, we provide a detailed description of the two stages, with the pipeline illustration in Appendix \ref{apx:algorithm}. 

\begin{figure*}[ht]
\centering
\includegraphics[width=\linewidth]{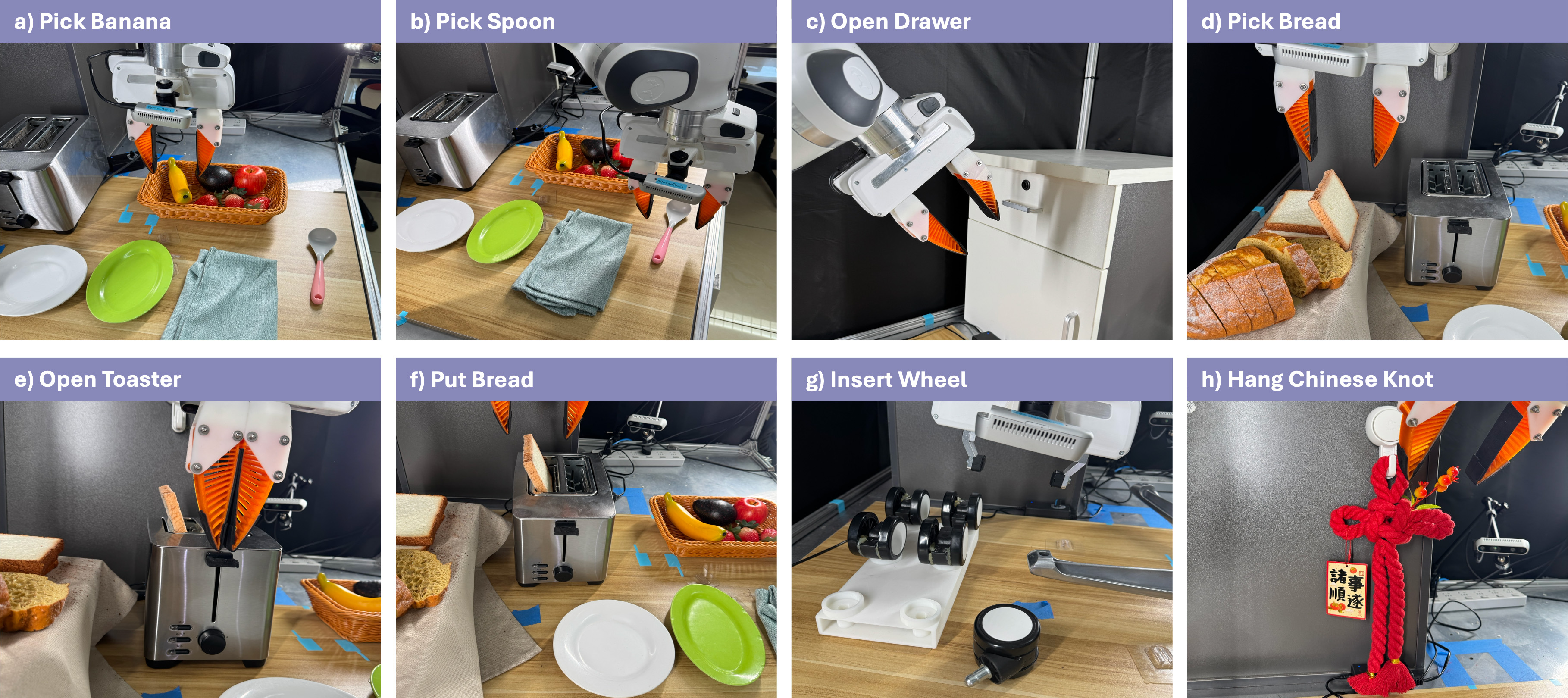}
\caption{\textbf{Overview of all real-world experimental tasks.} The real-world tasks include picking and placing (a) banana, (b) spoon, (d) and (f) bread, operating with (c) drawer and (e) toaster, assembling complex objects such as (g) chair wheel and (h) Chinese Knot. }
\label{fig:tasks}
\end{figure*}

\subsection{Stage I: Offline Fine-tuning with Cal-ConRFT}
Since pre-trained VLA models often lack zero-shot generalizability to novel robotic configurations, in the offline stage, we focus on training the policy using a small, pre-collected offline dataset (20–30 demonstrations) before transitioning to online reinforcement learning.
We initialize the policy with the pre-trained VLA model for reinforcement learning, reducing both the exploration burden and the overall online training time. Considering the ability to utilize offline data effectively, we choose the Calibrated Q-Learning (Cal-QL) \cite{nakamoto2024cal} as our base offline RL method since we want the Q-function to be robust to out-of-distribution (OOD) actions. Specifically, Cal-QL trains the Q-function on a pre-collected dataset by reducing temporal difference (TD) error and an additional regularizer. This regularizer penalizes Q-values for OOD actions when they exceed the value of the reference policy $V^{\mu}(s)$, while compensating for this penalization on actions observed within the offline dataset. The Cal-QL training objective for the critic is given by:
\begin{equation}
    \begin{aligned}
    \mathcal{L}_{Q}^{offline}(\theta)=&\alpha(\mathbb{E}_{s\sim\mathcal{D},a\sim\pi(\cdot|s)}[\max(Q_{\theta}(s,a),V^{\mu}(s))]\\
    &-\mathbb{E}_{s,a\sim\mathcal{D}}[Q_{\theta}(s,a)])\\
    &+\frac{1}{2}\mathbb{E}_{(s,a,s')\sim\mathcal{D}}[(Q_{\theta}(s,a)-\mathcal{B}^{\pi}\overline{Q}_{\overline{\theta}}(s,a))^2] \\
    \end{aligned}
    \label{eq:calql_offline}
\end{equation}
\noindent where $Q_{\theta}$ is the learned Q-function parameterized by $\theta$, $\overline{Q}_{\overline{\theta}}$ is the delayed target Q-function parameterized by $\overline{\theta}$. $\mathcal{B}^{\pi}\overline{Q}(s,a)=r(s,a)+\gamma\mathbb{E}_{a'\sim\pi(\cdot|s')}(\overline{Q}(s',a'))$ is the Bellman backup operator. $\alpha$ is a hyper-parameter to control the conservative penalty. And $\mathcal{D}$ is the demo buffer that stores demonstrations.

However, while Cal-QL is generally efficient at leveraging offline datasets, it struggles to train an effective policy when only small set of demonstrations (e.g., 20–30) are available. In such cases, limited state coverage leads to poor value estimates, making it difficult for the policy to generalize to unseen states. By contrast, typical offline RL datasets are often collected from multiple behavior policies, providing a broad range of state coverage to reduce the distribution shift. Lacking this breadth, the Cal-QL loss alone may not adequately guide the learning process, resulting in poor performance.

To address this issue, we propose augmenting the offline training process by incorporating a BC loss. The BC loss directly minimizes the difference between the actions generated by the policy and those from the demonstrations. By incorporating BC loss, we encourage the model to imitate the behaviors from the demonstrations, providing additional supervisory signals during the offline stage. This helps the VLA model to learn a more effective policy and initialize a stable Q function with few demonstrations, especially in the case of contact-rich manipulation tasks where control precision is critical. 

Motivated by combining the BC loss with Q guidance under a consistency-based objective \cite{chen2023boosting}, we introduce Cal-ConRFT in the offline stage. This approach employs a consistency policy as the action head for fine-tuning the VLA model, addressing two key concerns: 1) it helps leverage inconsistent and sub-optimal demonstrations that often arise in pre-collected data, and 2) compared to diffusion-based action head, the consistency-based action head remains computationally lightweight for efficient inference \cite{chen2023boosting, xing2025consistency, prasad2024consistency}. The consistency policy is a diffusion-model-based policy \cite{li2024stabilizing} that learns to map random actions sampled from the unit Gaussian to generate actions drawn from the expert action distribution conditioned on the current state. For the consistency policy, we discretize the diffusion horizon $[\epsilon,K]$ into $M$ sub-intervals with boundaries $k_1=\epsilon \le k_2 \le \cdots \le k_m = K$ and $\epsilon = 0.002$. Specifically, the VLA model with a consistency policy as the action head is given by:
\begin{equation}
    \begin{aligned}
    \pi_{\psi}(a|s) &= f_{\psi}(a^k,k|E_{\phi}(s))\\
    \end{aligned}
    \label{eq:cp}
\end{equation}
\noindent where $f$ denotes the consistency policy parameterized with $\psi$, subscripts $k$ denoted the diffusion step, $a^k\sim\mathcal{N}(0,kI)$ and $E_{\phi}(s)$ denotes the encoded state of the pre-trained VLA model parameterized with $\phi$. The consistency-based training objective for VLA model fine-tuning is given by:
\begin{equation}
    \begin{aligned}
    &\mathcal{L}_{\pi}^{offline}(\psi) = \beta\mathcal{L}_{\pi}^{BC} + \eta\mathcal{L}_{\pi}^{Q}\\
    \end{aligned}
    \label{eq:cpql_offline}
\end{equation}

\noindent where BC loss $\mathcal{L}_{\pi}^{BC}=\mathbb{E}_{(s,a)\sim\mathcal{D},m\sim\mathcal{U}[1,M-1]}[d(f_{\psi}(a+k_mz,k_m|E(s)),a)]$, $z\sim\mathcal{N}(0,I)$, $d$ stands for the Euclidean distance $d(x,y)=||x-y||_2$, and Q loss $\mathcal{L}_{\pi}^{Q}=-\mathbb{E}_{s\sim\mathcal{D},a\sim\pi_{\psi}}[Q(s,a)]$. $\beta$ and $\eta$ are two hyper-parameters to balance the BC loss and Q loss. This combination enables efficient policy learning and stable value estimation, even with a small set of demonstrations, by aligning value estimates with expert actions and improving policy performance during offline training. Moreover, it provides a reliable initialization for the online stage, facilitating safe and effective exploration.

\begin{table*}
    \centering
    \scalebox{0.87}{
    \begin{tabular}{r|c|ccccc|ccccc}
    &\multirow{2}{*}{\makecell[c]{\textbf{Training}\\\textbf{Time}\\\textbf{(mins)}}} &\multicolumn{5}{c}{\textbf{Success Rate (\%)}} &\multicolumn{5}{c}{\textbf{Episode length}}  \\
    \textbf{Task} & &\makecell[c]{SFT\cite{team2024octo}} &\makecell[c]{Cal-\\ConRFT} &\makecell[c]{HG-\\DAgger\cite{kelly2019interactive}} &PA-RL\cite{mark2024policy} &\makecell[c]{HIL-\\ConRFT} &\makecell[c]{SFT\cite{team2024octo}} &\makecell[c]{Cal-\\ConRFT} &\makecell[c]{HG-\\DAgger\cite{kelly2019interactive}} &PA-RL\cite{mark2024policy} &\makecell[c]{HIL-\\ConRFT} \\ 
    \hline
    Pick Banana       &45   &40   &50   &60 (+50\%)   &80 (+100\%)  &\textbf{90} (+80\%)    &63.7 &57.8 &67.5 (0.9x) &56.1 (1.1x) &\textbf{51.2} (1.1x) \\
    Put Spoon         &45   &50   &55   &90 (+80\%)   &80 (+60\%)   &\textbf{100} (+82\%)   &49.9 &57.2 &50.9 (1.0x) &45.3 (1.1x) &\textbf{22.6} (2.5x) \\
    Open Drawer       &15   &35   &30   &80 (+129\%)  &60 (+71\%)   &\textbf{100} (+233\%)  &63.6 &61.7 &48.4 (1.3x) &57.1 (1.1x) &\textbf{32.4} (1.8x) \\
    Pick Bread        &45   &65   &55   &65 (+0\%)    &80 (+23\%)   &\textbf{100} (+82\%)   &53.2 &49.1 &65.6 (0.8x) &51.7 (1.0x) &\textbf{31.6} (1.6x) \\
    Open Toaster      &30   &30   &30   &75 (+116\%)  &100 (+233\%) &\textbf{100} (+233\%)  &51.2 &50.7 &43.4 (1.2x) &34.3 (1.5x) &\textbf{22.1} (2.3x) \\
    Put Bread         &60   &5    &20   &60 (+1100\%) &75 (+1400\%) &\textbf{100} (+400\%)  &102  &84.8 &74.2 (1.4x) &72.1 (1.4x) &\textbf{36.6} (2.3x) \\
    Insert Wheel      &60   &35   &35   &40 (+14\%)   &30 (-14\%)   &\textbf{80} (+129\%)   &42.7 &43.4 &53.0 (0.8x) &47.4 (0.9x) &\textbf{21.9} (2.0x) \\
    Hang Chinese Knot &90   &55   &40   &50 (-10\%)   &65 (+18\%)   &\textbf{100} (+150\%)  &52.6 &54.9 &47.5 (1.1x) &44.4 (1.3x) &\textbf{26.8} (2.0x) \\
    \hline
    \rowcolor{gray!20} \textbf{Average}  &48.8 &39.4 &39.4 &65 (+65\%) &71.3 (+81\%) &\textbf{96.3} (+144\%) &59.9 &57.5 &56.3 (1.1x) &51.1 (1.2x) &\textbf{30.7} (1.9x) \\
    \end{tabular}}
    \caption{\textbf{All experiment results for various offline and online fine-tuning methods.} We report the policy performance against various baselines after offline fine-tuning (SFT \cite{team2024octo} and Cal-ConRFT) and after online fine-tuning (HG-DAgger \cite{kelly2019interactive}, PA-RL \cite{mark2024policy} and HIL-ConRFT), including success rates and average episode lengths for various tasks. Specifically, for online fine-tuning, HG-DAgger, and PA-RL training starts from the SFT baseline, and HIL-ConRFT training starts from the Cal-ConRFT baseline. The performance improvement is relative to corresponding offline results. Policies are trained using the same number of online episodes with human interventions for all methods. All metrics are reported over 20 trials per task. }
    \label{tab:exp_result}
\end{table*}

\subsection{Stage II: Online Fine-tuning with HIL-ConRFT}

While the offline stage provides an initial policy from a small set of demonstration data, its performance is limited by the scope and quality of the pre-collected demonstrations. Therefore, we have the online stage with HIL-ConRFT, where the VLA model is further fine-tuned online via the consistency policy through interacting with the real-world environment. During online training process, the demo buffer $\mathcal{D}$ for offline stage is remained. Furthermore, we have a replay buffer $\mathcal{R}$ to store online data, then implement symmetric sampling \cite{ball2023efficient}, whereby for each batch, we sample equally between these two buffers to form each training batch. Since the VLA model continuously gathers new transitions based on its current policy, the data distribution naturally evolves with the policy. This ongoing interaction reduces the distribution-shift problem that the offline stage faces. As a result, we use a standard Q loss for online critic updating: 
\begin{equation}
    \begin{aligned}
    \mathcal{L}_{Q}^{online}(\theta)&=\mathbb{E}_{(s,a,s')\sim(\mathcal{D}\cup\mathcal{R})}[(Q_{\theta}(s,a)-\mathcal{B}^{\pi}\overline{Q}(s,a))^2] \\
    \end{aligned}
    \label{eq:ql_online}
\end{equation}
The consistency-based training objective for VLA model fine-tuning is given by:
\begin{equation}
    \begin{aligned}
    &\mathcal{L}_{\pi}^{online}(\psi) = \beta\mathcal{L}_{\pi}^{BC} + \eta\mathcal{L}_{\pi}^{Q}
    \end{aligned}
    \label{eq:cpql_online}
\end{equation}
\noindent where BC loss $\mathcal{L}_{\pi}^{BC}=\mathbb{E}_{(s,a)\sim(\mathcal{D}\cup\mathcal{R}),m\sim\mathcal{U}[1,M-1]}[d(f_{\psi}(a+k_mz,k_m|E(s)),a)]$, $z\sim\mathcal{N}(0,I)$, $d$ stands for the Euclidean distance $d(x,y)=||x-y||_2$, and Q loss $\mathcal{L}_{\pi}^{Q}=-\mathbb{E}_{s\sim(\mathcal{D}\cup\mathcal{R}),a\sim\pi_{\psi}}[Q(s,a)]$. Note that this objective closely mirrors Equation \ref{eq:cpql_offline} from the offline stage, enabling a quick adaption to online fine-tuning.

Typically, we decrease the BC loss weight $\beta$ while increasing the Q loss weight $\eta$ during the online stage, yet we keep the BC loss for two main reasons. 1) Firstly, it ensures the policy continues to align with the demonstration data, preventing drastic deviations that could lead to performance collapse. This is important for maintaining the quality of actions in contact-rich manipulation tasks, where sudden changes in the policy can result in unsafe or inefficient behaviors. 2) Secondly, since reinforcement learning inherently involves exploration, it can become unstable in high-dimensional state-action spaces. By providing a stabilizing effect on exploration \cite{li2024generalizing}, the BC loss prevents the policy from deviating too far from its offline baseline, thereby reducing the risk of inefficient or unsafe behaviors. This aspect is important in real-world robotic training, especially in physical environments where unsafe actions can lead to damage or other hazards.

Also, we integrate human interventions into the online stage through Human-in-the-Loop learning. Specifically, HIL learning allows for timely interventions by a human operator who can provide corrective actions during the exploration process, which will then take over the control of the robot from the VLA model. These human corrections are added to the demo buffer $\mathcal{D}$, offering high-level guidance that steers exploration in a safer and more efficient direction \cite{huihan2023robot}. Human interventions are essential when the robot engages in destructive behaviors, such as colliding with obstacles, applying excessive force, or damaging the environment. In addition to ensuring safe exploration, human interventions accelerate policy convergence. In scenarios where the policy leads the robot into an unrecoverable or undesirable state or when the robot becomes stuck in a local optimum that would otherwise require significant time and steps to overcome without external assistance, the human operator can step in to correct the robot’s actions and guide it towards safer and more effective behavior. This results in a stable learning process, where the VLA model is fine-tuned quicker and more safely than it would through autonomous exploration alone. 

\section{Experiment and Results}

In this section, we validate the proposed fine-tuning framework through real-world experiments. We first present the experimental setup and the results for various baselines and then discuss these results and their implications.

\begin{figure*}[ht]
\centering
\includegraphics[width=\linewidth]{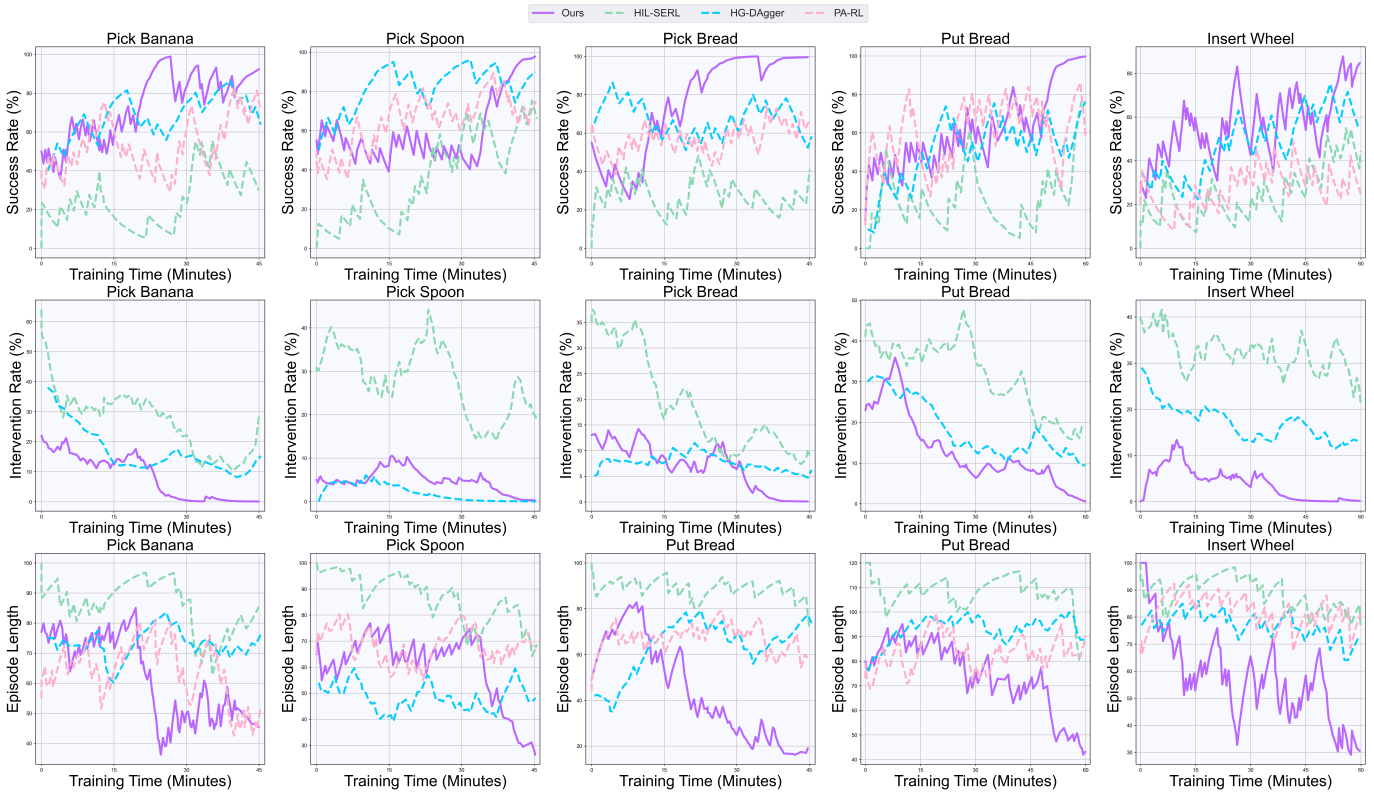}
\caption{\textbf{Learning curves during online training.} This figure presents the success rates, intervention rates, and episode lengths for HIL-SERL \cite{luo2024precise}, HG-DAgger \cite{kelly2019interactive}, PA-RL \cite{mark2024policy} and our method across five representative real-world tasks, displayed as a running average over 20 episodes. PA-RL is implemented without human intervention. Note that human interventions may lead the policy to successful outcomes, and thus, the actual policy success rate when interventions exist might be lower than the curve suggests. }
\label{fig:online_result}
\end{figure*}

\subsection{Overview of Experiments}

Our experiments aim to evaluate our approach's effectiveness and efficiency for fine-tuning VLA models in real-world scenarios. To this end, we perform real-world experiments across eight diverse manipulation tasks, as illustrated in Figure \ref{fig:tasks}. These tasks are designed to reflect a variety of manipulation challenges, including object placement tasks (e.g., placing bread into a toaster and putting bread on a white plate), precise and contact-rich manipulation (e.g., aligning and inserting a wheel into the chair base), and dynamic object handling (e.g., hanging a Chinese Knot). To validate our fine-tuning approach, we select the Octo-small model \cite{team2024octo} for its balance of performance and inference efficiency, and employ a consistency policy \cite{prasad2024consistency} as the action head on a 7-DoF Franka Emika robot arm.

For all tasks, the state observation includes two RGB images captured from a wrist-mounted camera (128 × 128) and a side camera (256 × 256), in combination with the robot's proprioceptive state of the robot arm, including end-effector poses, twists, forces/torques, and gripper status. The action space is defined as either a 6-dimensional end-effector delta pose for the downstream impedance controller or a 7-dimensional target that includes 1-dimensional binary gripper action, additionally for tasks that involve grasping. Data collection and policies command actions at 10Hz. Before training, positive and negative demonstrations are collected from human operators to train a binary classifier that gives binary feedback on whether the corresponding task is done successfully or not for each task. Additionally, each task's initial state is randomized using either a scripted robot motion or manual resets by a human operator. We present descriptions of each task in our real-world experiments and more details on the experiment tasks, training, and evaluation procedure in the Appendix \ref{apx:tasks}.

\subsection{Experimental Results}

In this section, we provide the experimental results for all tasks as shown in Figure \ref{fig:tasks}. For each task, we report result metrics, including the success rate, episode length, and total training time in Table \ref{tab:exp_result}. The training time includes the duration of scripted motions, policy rollouts, and onboard computations, all of which are conducted using an NVIDIA RTX A6000 GPU. For the offline stage, we compare Cal-ConRFT and SFT, where the SFT uses an NLL loss for behavior cloning \cite{team2024octo}. For the online stage, we compared HIL-ConRFT with multiple baselines, including HG-DAgger \cite{kelly2019interactive} that incorporates human corrections to fine-tune the policy through supervised learning, PA-RL \cite{mark2024policy} that optimized actions through a policy-agnostic Q-function and fine-tune the policy through supervised learning with the optimized actions. We also compare HIL-SERL \cite{luo2024precise} that trains a RL policy with human interventions from scratch, and RLDG \cite{xu2024rldg} that fine-tunes the VLA model using SFT \cite{team2024octo} with demonstrations collected by RL policy. 

\subsubsection{ConRFT Outperforms Supervised Methods}

We compare different approaches for supervised and reinforced methods in Table \ref{tab:exp_result} and present the corresponding online learning curves in Figure \ref{fig:online_result}. Our approach, ConRFT, achieves the highest average success rate of 96.3\% after 45 to 90 minutes of real-world training across all tasks, representing a 144\% improvement over the supervised baseline. It outperforms SOTA methods such as HG-DAgger and PA-RL, with average success rates of 65\% and 71.3\%. While HG-DAgger leverages human corrections to fine-tune the VLA model through supervised learning, it fails to achieve significant policy improvement and even experiences a performance drop on some tasks due to the sub-optimality and inconsistency of human corrections. For example, we observe that for contact-rich tasks that require precise, careful manipulation, such as Insert Wheel and Hang Chinese Knot, HG-DAgger has limited policy improvement after online fine-tuning. Specifically, in the Hang Chinese Knot task, the careful handling of soft objects demands consistent and precise controls. The inherent variability in human corrections, such as differences in the angle of insertion, introduces noise and conflicting information into the training process. This inconsistency prevents the policy's ability to learn precise and dexterous behaviors. Additionally, the complexity of contact dynamics means that minor deviations in the policy can result in significant performance drops, further exacerbating the challenges posed by inconsistent human corrections. 

In the absence of human corrections, PA-RL offers a direct action optimization using a policy-agnostic Q-function trained through Cal-QL. By optimizing actions based on reward signals, PA-RL overcomes the sub-optimality of human corrections and demonstrates more stable policy improvement in simpler tasks such as Pick Banana and Put Spoon. However, it fails to improve the policy performance in contact-rich tasks that require precise, careful manipulation, such as Insert Wheel. Precise alignment and controlled insertion forces are critical in the Insert Wheel task. However, due to the limited state coverage in the demo buffer and replay buffer, the policy-agnostic Q-function is unable to generalize effectively to different wheel and slot positions. This limits the policy's ability to handle the slight variations in state transitions required for successful insertion, leading to sub-optimal performance in complex manipulation scenarios. Consequently, while PA-RL shows promise in simple environments, it struggles to scale to complex tasks demanding high precision and dexterity.

These observations underscore the advantages of our proposed approach, which effectively mitigates the issues associated with inconsistent human corrections and limited state coverage by reinforcement learning. Our method, ConRFT, effectively and safely explores a broad range of states and directly optimizes the policy using task-specific rewards, thereby demonstrating high sample efficiency and mitigating the impact of inconsistent human corrections. This stability and performance highlight the effectiveness of our approach in overcoming the limitations of existing fine-tuning methods in real-world robotic applications.

\begin{table}[t]
    \centering
    \scalebox{0.82}{
    \begin{tabular}{r|c|cc|cc}
         &\multirow{2}{*}{\makecell[c]{\textbf{Training}\\\textbf{Time}\\\textbf{(mins)}}} &\multicolumn{2}{c}{\textbf{Success Rate (\%)}} &\multicolumn{2}{c}{\textbf{Episode length}}  \\
        \multirow{2}{*}{\textbf{Task}} & &\makecell[c]{HIL-\\SERL\cite{luo2024precise}} &\makecell[c]{HIL-\\ConRFT} &\makecell[c]{HIL-\\SERL\cite{luo2024precise}} &\makecell[c]{HIL-\\ConRFT} \\ 
        \hline
        Pick Banana       &45   &0 $\rightarrow$ 15           &50 $\rightarrow$ \textbf{90}  &\textbf{30.6} &51.2 \\
        Put Spoon         &45   &0 $\rightarrow$ 60           &55 $\rightarrow$ \textbf{100} &56.1          &\textbf{22.6} \\
        Open Drawer       &15   &0 $\rightarrow$ 10           &30 $\rightarrow$ \textbf{100} &67.5          &\textbf{32.4} \\
        Pick Bread        &45   &0 $\rightarrow$ 45           &55 $\rightarrow$ \textbf{100} &\textbf{22.0} &31.6 \\
        Open Toaster      &30   &0 $\rightarrow$ \textbf{100} &30 $\rightarrow$ \textbf{100} &28.1          &\textbf{22.1} \\
        Put Bread         &60   &0 $\rightarrow$ 5            &20 $\rightarrow$ \textbf{100} &62.0          &\textbf{36.6} \\
        Insert Wheel      &60   &0 $\rightarrow$ 5            &35 $\rightarrow$ \textbf{80}  &42.0          &\textbf{21.9} \\
        Hang Chinese Knot &90   &0 $\rightarrow$ 15           &40 $\rightarrow$ \textbf{100} &57.3          &\textbf{26.8} \\
        \hline
        \rowcolor{gray!20} \textbf{Average}  &48.8 &0 $\rightarrow$ 31.9 &39.4 $\rightarrow$ \textbf{96.3} &45.7 &\textbf{30.7} \\
    \end{tabular}}
    \caption{\textbf{Experiment results for training from scratch (HIL-SERL \cite{luo2024precise}) and fine-tuning VLA (HIL-ConRFT).} Policies are trained using the same number of episodes with human interventions. All metrics are reported over 20 trials per task.}
    \label{tab:offline_rl_result}
\end{table}

Another critical metric for evaluating policy performance is the episode length, which represents the total number of steps the policy takes to complete a task. As shown in Table \ref{tab:exp_result}, the VLA model fine-tuned with HIL-ConRFT achieves an average episode length of 30.7 steps, demonstrating a 1.9x shorter than the offline baselines. In contrast, HG-DAgger achieves an average episode length of 56.3 steps, which is only 1.1x shorter than the offline baseline. Similarly, PA-RL attains an average episode length of 51.1 steps. It lacks policy exploration due to the conservative nature of the policy-agnostic Q-function, preventing it from effectively optimizing how to complete the task more quickly or trying more efficient behaviors. 

These results illustrate that ConRFT can effectively exploit the dynamic characteristics of MDPs to optimize the VLA model via consistency policy for maximizing the discounted sum of rewards. They also show the limitations of supervised methods in handling sub-optimal data and efficient policy exploration. By encouraging policies to obtain rewards more quickly, our approach results in shorter episode lengths than supervised methods relying solely on imitating demonstrations. This enhanced sample efficiency and reduced episode length highlight the advantages of ConRFT for fine-tuning VLA models in real-world robotic applications.

\subsubsection{Fine-tuning VLA Outperforms Training From Scratch}

Reinforcement learning from scratch typically demands extensive interaction with the environment and frequent human interventions, which can lead to a lengthy training process and high safety risks. For instance, HIL-SERL \cite{luo2024precise}, an approach that trains policies through RL from scratch with human interventions, fails to converge to an effective policy within the same training duration as our approach, reaching an average success rate of only 31.9\% as shown in Table \ref{tab:offline_rl_result}. The learning curves in Figure \ref{fig:online_result} reveal that HIL-ConRFT consistently improves policy performance during the online stage. While HIL-SERL can achieve optimal policies eventually, it usually requires over two hours of online training with a higher intervention rate for each task, resulting in more destructive behaviors during exploring (e.g., collisions with the environment), especially in the early stage of training.

In contrast, starting from a pre-trained VLA model and performing offline fine-tuning reduces the online training time and improves sample efficiency. Building upon offline initialized policy, ConRFT accelerates the policy convergence and enhances the final performance. As a result, fine-tuning VLA models via consistency policy enables them to reach higher success rates more quickly and with fewer interventions compared to training entirely from scratch, demonstrating the benefits of leveraging pre-trained VLA models in real-world robotic applications.

\begin{figure}[t]
\centering
\includegraphics[width=\linewidth]{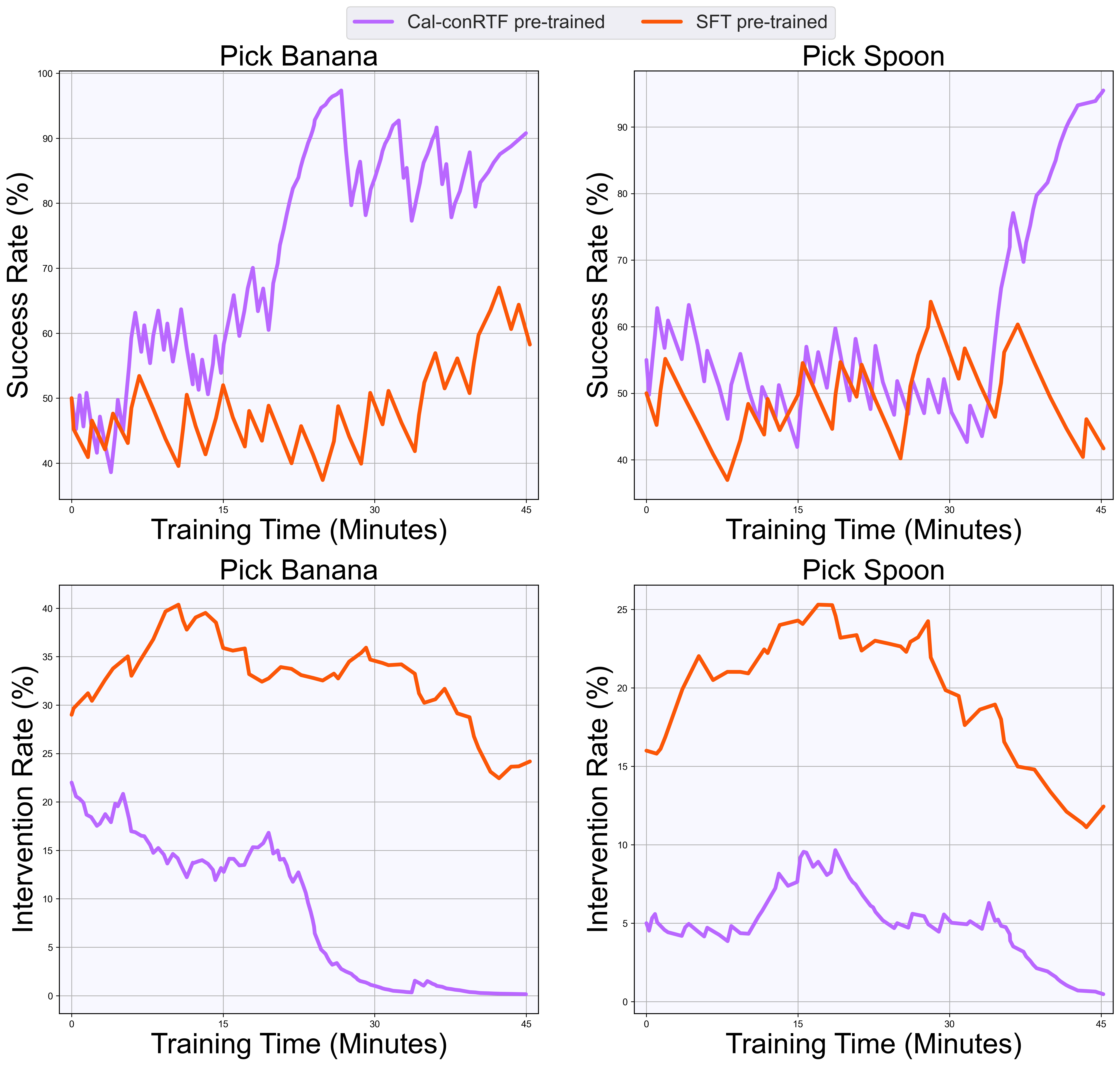}
\caption{\textbf{Learning curves for HIL-ConRFT online fine-tuning from SFT \cite{team2024octo} and Cal-ConRFT baselines.} This figure presents success and intervention rates across two representative tasks, displayed as a running average over 20 episodes. }
\label{fig:abla_bc}
\end{figure}

\subsubsection{Analysis}

\paragraph{Why fine-tuning from Cal-ConRFT rather than SFT or Cal-QL?}

As illustrated in Table \ref{tab:exp_result}, we observe that during the offline stage, the performance of Cal-ConRFT is similar to that of the SFT baseline. This observation raises the question of why Q loss should be introduced during the offline stage. The reason is that when the offline stage relies solely on SFT, the fine-tuned policy benefits from imitation learning but may require substantial online fine-tuning to handle states and actions not covered by the offline dataset. In contrast, incorporating Q loss during the offline stage allows the early Q-value estimations to provide initialization for policy improvement, facilitating quicker adaptation during online fine-tuning. This approach helps address potential biases and ensures more stable learning. Moreover, in scenarios with a small set of demonstrations, we find that relying on Cal-QL alone is insufficient to train an effective policy, resulting in a 0\% success rate on all tasks. The lack of data affects the policy's ability to accurately estimate Q-values, resulting in weak performance after the offline stage and longer training time in the online stage. 

We compare the online fine-tuning curves starting from Cal-ConRFT and the SFT baseline on two representative tasks to investigate further the impact of introducing Q loss, as shown in Figure \ref{fig:abla_bc}. Although both curves begin with similar success rates, the higher intervention rate observed during training from the SFT baseline indicates that the SFT-trained policy experiences severe policy forgetting in the early stages of online training. This suggests that Cal-ConRFT enables quicker adaptation of the online learning process by leveraging the Q loss during the offline stage, allowing more effective and stable policy improvement with a small set of demonstration data. 



\begin{table}[t]
    \centering
    \scalebox{0.88}{
    \begin{tabular}{r|ccccc}
          &\multicolumn{5}{c}{\textbf{Success Rate (\%)}}\\
        \textbf{Task}     &DP\cite{chi2023diffusion} &SFT\cite{team2024octo} &RLDG\cite{xu2024rldg} &Cal-ConRFT &HIL-ConRFT\\ 
        \hline
        Put Spoon         &60   &70   &\textbf{100} &55         &\textbf{100} \\
        Put Bread         &30   &65   &\textbf{100} &20         &\textbf{100} \\
        Insert Wheel      &35   &40   &50           &35         &\textbf{80}  \\
        \hline
        \rowcolor{gray!20} \textbf{Average} &41.7 &58.3 &83.3 &36.7 &\textbf{93.3} \\
    \end{tabular}}
    \caption{\textbf{Experimental comparisons with various demonstrations.} Diffusion Policy (DP) \cite{chi2023diffusion} and SFT \cite{team2024octo} are trained with 150 demonstrations collected by human teleoperation, while RLDG \cite{xu2024rldg} is trained with 150 demonstrations collected by RL policy. Cal-ConRFT is trained with 20 demonstrations collected by human teleoperation, and HIL-ConRFT is trained with 20 demonstrations as well as 80-120 policy-generated rollout trajectories. All metrics are reported over 20 trials per task.}
    \label{tab:offline_bc_result}
\end{table}

\paragraph{Does increasing the number of demonstrations enhance policy performance for SFT?}

Typically, during a 45-60 minutes online fine-tuning stage, the policy collects approximately 80 to 120 successful and failed trajectories. To ensure a fair comparison between our approach and supervised training methods, we further compare training Diffusion Policy (DP) \cite{chi2023diffusion} and supervised fine-tuning VLA \cite{team2024octo} using 150 demonstrations on three representative tasks, which aligns with the total number of demonstrations utilized by our approach. Additionally, we compare RLDG \cite{xu2024rldg} with fine-tuning using 150 demonstrations collected by RL policy. As shown in Table \ref{tab:offline_bc_result}, even though DP and SFT benefit from a larger quantity of demonstrations, their success rates still fail to match the performance of our method, especially on contact-rich tasks such as Insert Wheel. This indicates that simply adding more human-collected demonstrations with supervised learning does not necessarily guarantee higher performance due to the inconsistent and sub-optimal actions inherent in human-collected data. Meanwhile, RLDG achieves higher success rates using optimal data collected from RL policies, suggesting that the consistency of these RL-collected data can improve the final performance. On the other hand, our method directly fine-tune the policy by optimizing the consistency-based training objective, achieving the highest success rate.

\begin{table}[t]
\centering
\scalebox{1.0}{
\begin{tabular}{r|cc}
     &\multicolumn{2}{c}{\textbf{Success Rate (\%)}} \\
    \textbf{Task}     &Kosmos-2(1.6B)     &PaliGemma(3B) \\
    \hline
    Pick Banana       &60$\rightarrow$100 &65$\rightarrow$100 \\
    Put Spoon         &55$\rightarrow$100 &30$\rightarrow$100 \\
    Hang Chinese Knot &45$\rightarrow$100 &60$\rightarrow$100 \\
    \hline
    \rowcolor{gray!20} \textbf{Average} &53.3$\rightarrow$100 &51.7$\rightarrow$100 \\
\end{tabular}
}
\caption{\textbf{Experimental results of ConRFT on different VLA models.} We fine-tune RoboVLM \cite{li2024towards} with two VLM backbones using our method. Specifically, we fine-tune only the action head while keeping the visual encoders and transformer backbone frozen. All metrics are reported over 20 trials per task.}
\label{tab:more_vlas}
\end{table}

\paragraph{Practicality of ConRFT across Various VLA Models}

ConRFT is highly versatile and can be applied to any VLM-based architecture with an action head. This flexibility stems from its ability to optimize the action generation process independently of the underlying visual encoder, making it adaptable to various VLA frameworks. To further validate its applicability generalization, we test out approach on fine-tuning RoboVLM \cite{li2024towards} with two distinct VLM backbones. As shown in Table \ref{tab:more_vlas}, the results indicate that ConRFT can effectively enhance the performance of various VLAs, improving the success rates across multiple robotic tasks. This ability to fine-tune the action generation while leveraging the pretrained visual components underscores the broad applicability of ConRFT.

\section{Limitations}

Although our approach demonstrates strong performance and sample efficiency for fine-tuning VLA models in real-world manipulation tasks, several limitations remain.

\subsection{Sensitivity to Reward Engineering}

In this work, we implement a task-specific binary classifier to calculate the reward for RL. However, the inherent distributional shift between the classifier's training data and the state-action distributions generated during RL exploration creates a critical vulnerability, as it can lead the learned policy to engage in reward hacking, exploiting unintended behaviors where the classifier provides inaccurate rewards. For instance, the robot might position its end-effector at a specific location that triggers a false positive, causing the policy to converge to an incorrect behavior. Since these reward classifiers typically provide only sparse feedback, the policy may learn slowly, even with the help of human interventions. On the other hand, this reward-driven approach leads to highly specialized policies that are closely tied to the specific conditions of the task, limiting their ability to generalize to new environments. While introducing multi-task dense reward signals could improve sample efficiency and accelerate policy convergence, it would also demand more sophisticated reward engineering for real-world applications.

\subsection{Frozen Encoders and Transformer Backbone}

Our current implementation runs the interaction and policy learning processes in separate threads, fine-tuning only the action head network with consistent policy while keeping the visual encoders and transformer backbone frozen. While this design choice boosts real-time performance, it constrains the policy’s ability to refine its perception and representation modules during online training, especially for unseen scenarios. Allowing partial or complete updates of these frozen components, potentially with efficient techniques such as parameter-efficient fine-tuning (e.g., LoRA \cite{hu2021lora}), could enhance final task performance and adaptability without sacrificing safety or speed.

\section{Conclusion} 

We presented a two-stage approach, ConRFT, for reinforced fine-tuning VLA models in real-world robotic applications. By first performing offline fine-tuning (Cal-ConRFT) with a small set of demonstrations, we initialize a reliable policy and value function via a unified training objective that integrates Q loss and BC loss in a consistency-based framework. We then leveraged task-specific rewards and human interventions in the online fine-tuning stage (HIL-ConRFT) to fine-tune the VLA model via consistency policy. Experiments on eight diverse real-world tasks demonstrated that our approach outperforms SOTA methods in terms of success rate, sample efficiency, and episode length. Overall, this work showcases a practical way to use reinforcement learning for safe and efficient VLA model fine-tuning.

\section*{Acknowledgments}
This work is supported by the National Natural Science Foundation of China (NSFC) under Grant No. 62136008 and in part by the International Partnership Program of the Chinese Academy of Sciences under Grant 104GJHZ2022013GC.

\bibliographystyle{unsrtnat}

\clearpage
\appendix

\subsection{Algorithm Illustration}
\label{apx:algorithm}
The whole pipeline of conRFT is outlined in Algorithm \ref{alg:conrft}.

\begin{algorithm}[ht]
    \caption{Procedure of ConRFT } 
    \begin{algorithmic}
        \REQUIRE A pre-trained VLA model $\pi_{\theta, \psi}$ with VLA parameter $\phi$ and a consistency head parameter $\psi$. A critic model $Q$ with parameter $\theta$. A pre-collected dataset $\mathcal{D}$ including 20-30 demonstrations. 
        Initialize batch size $B$.
        \STATE Randomly initialize the action head $\psi$ and the critic model $\theta$
        \STATE \textcolor{gray!90}{\# Stage I: Offline fine-tuning with Cal-ConRFT}
        \FOR{each offline training step}
            \STATE Sample $(s_t, a_t, r_t, s_{t+1})$ of $batch\_size$ from $\mathcal{D}$
            \STATE Update the action head $\psi$ and the critic model $\theta$ by Equation \ref{eq:calql_offline} and Equation \ref{eq:cpql_offline}.
        \ENDFOR
        \STATE \textcolor{gray!90}{\# Stage II: Online fine-tuning with HIL-ConRFT}
        \STATE \textcolor[RGB]{18,220,168}{\textbf{\# Start Policy Learning Thread}:}
        \STATE \textbf{Wait until} The number of transitions in $\mathcal{R}$ is at least 100
        \FOR{each online training step}
            \STATE Sample $(s_t, a_t, r_t, s_{t+1})$ of $\frac{B}{2}$ from $\mathcal{D}$ and $\mathcal{R}$
            \STATE Combine both minibatches to form batch of size $B$
            \STATE Update the action head $\psi$ and the critic model $\theta$ by Equation \ref{eq:ql_online} and Equation \ref{eq:cpql_online}.
        \ENDFOR
        \STATE \textcolor[RGB]{26,205,230}{\textbf{\# Start Interaction Thread}:}
        \FOR{each interaction step}
            \IF{no human intervention} 
                \STATE Take action $a_t \sim \pi_{\psi}(\cdot|s_t)$
                \STATE Store transition $(s_t,a_t,r_t,s_{t+1})$ in $\mathcal{R}$
            \ELSE 
                \STATE Take action $a_{intv}$
                \STATE Store transition $(s_t,a_{intv},r_t,s_{t+1})$ in $\mathcal{D}$
            \ENDIF
        \ENDFOR
    \end{algorithmic} 
    \label{alg:conrft}
\end{algorithm}

\subsection{Task Description, Setup and Policy Training Details}

\label{apx:tasks}
In this section, we provide hardware setup and training details for each task. The 6-dimensional action space refers to the 6-dimensional end-effector delta pose, and the 7-dimensional action space includes the 6-dimensional end-effector delta pose and 1-dimensional gripper control action. The learning rate is 3e-4, and the batch size is 256 for all tasks. 

For the consistency policy utilized for fine-tuning VLA models, we set $k \in [0.002, 80.0]$ and the number of sub-intervals $M=40$ where the sub-interval boundaries are determined with the formula $k_i=(\epsilon^{\frac{1}{\rho}}+\frac{i-1}{M-1}(T^{\frac{1}{\rho}}-\epsilon^{\frac{1}{\rho}}))^{\rho}$, where $\rho=7$. The network is based on a 2-layer multi-layer perceptron (MLP), with a hidden size of 256 and the Mish function serving as the activation function. 

For the diffusion policy, we use diffusion steps $K=5$, a cosine beta schedule, Resnet 18, and the $ LN_Resnet$ architecture, with a hidden size of 256 and $n=3$ blocks.  

For the reward of all tasks, we give $+10$ reward when the task is completed and a $-0.05$ reward on each step. For HIL-SERL, which uses a DQN network for gripper control, we give a $-0.2$ reward every time the policy opens/closes the gripper. 

\paragraph{Pick Banana}

\begin{figure*}[htbp]
    \centering
    \includegraphics[width=\linewidth]{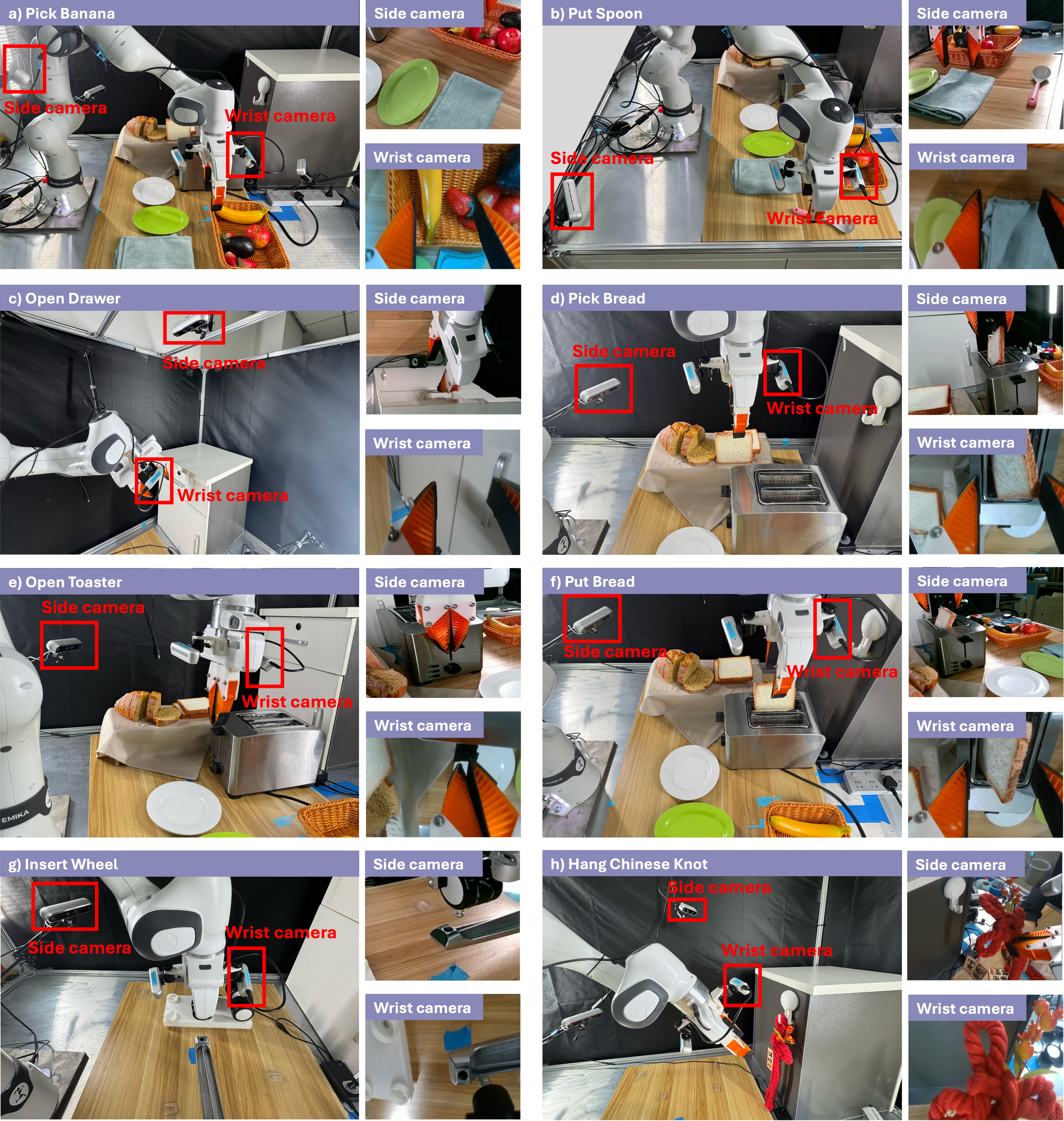}
    \caption{\textbf{Hardware setup and illustrations of camera views.} We give the illustrations of hardware setup and the corresponding camera views for all real-world tasks in this paper, including a) Pick Banana, b) Put Spoon, c) Open Drawer, d) Pick Bread, e) Open Toaster, f) Put Bread, g) Insert Wheel, h) Hand Chinese Knot.}
    \label{fig:tasks_detail}
\end{figure*}

This task involves picking up a banana in the basket and placing it on a green plate, which requires control of the gripper to move the fruit, as shown in Figure \ref{fig:tasks_detail}. It requires the policy to grasp and place the banana, ensuring it remains intact while avoiding collisions with the surrounding environment, such as the basket. We report more specific details of the policy training for this task in Table \ref{tab:task1_detail}. The task description for the VLA model is "Put the yellow banana on the green plate." 

\begin{table}[htbp]
    \centering
    \begin{tabular}{c|c}
        Parameter &Value\\
        \hline 
        Action space &7-dimensional\\
        Initial offline demonstrations &20\\
        Max episode length &100\\
        Reset method &Human reset\\
        Randomization range & 3 cm in x and y\\
        $(\alpha, \beta, \eta)$ for offline fine-tuning & $(0.01, 1.0, 0.1)$\\
        $(\beta, \eta)$ for online fine-tuning & $(0.5, 1.0)$\\
    \end{tabular}
    \caption{\textbf{Policy training details for the Pick Banana task.} }
    \label{tab:task1_detail}
\end{table}

\paragraph{Put Spoon}

This task involves picking up a spoon and placing it on a blue table linen, which requires the gripper to grasp and put the spoon, as shown in Figure \ref{fig:tasks_detail}. The challenge lies in the control needed to grasp the spoon. We report more specific details of the policy training for this task in Table \ref{tab:task2_detail}. The task description for the VLA model is "Put the spoon on the blue towel."

\begin{table}[htbp]
    \centering
    \begin{tabular}{c|c}
        Parameter &Value\\
        \hline 
        Action space &7-dimensional\\
        Initial offline demonstrations &20\\
        Max episode length &100\\
        Reset method &Human reset\\
        Randomization range & 3 cm in x and y\\
        $(\alpha, \beta, \eta)$ for offline fine-tuning & $(0.01, 1.0, 0.1)$\\
        $(\beta, \eta)$ for online fine-tuning & $(0.5, 1.0)$\\
    \end{tabular}
    \caption{\textbf{Policy training details for the Put Spoon task.} }
    \label{tab:task2_detail}
\end{table}

\paragraph{Open Drawer}

This task involves opening a drawer by grasping the handle and pulling it outward, as shown in Figure \ref{fig:tasks_detail}. It requires the policy to securely grip the handle and apply the correct force to open the drawer without damaging the hinges or surrounding area. We report more specific details of the policy training for this task in Table \ref{tab:task3_detail}. The task description for the VLA model is "Open the drawer."

\begin{table}[htbp]
    \centering
    \begin{tabular}{c|c}
        Parameter &Value\\
        \hline 
        Action space &6-dimensional\\
        Initial offline demonstrations &20\\
        Max episode length &100\\
        Reset method &Script reset\\
        Randomization range & 3 cm in y and x\\
        $(\alpha, \beta, \eta)$ for offline fine-tuning & $(0.01, 1.0, 0.1)$\\
        $(\beta, \eta)$ for online fine-tuning & $(0.5, 1.0)$\\
    \end{tabular}
    \caption{\textbf{Policy training details for the Open Drawer task.} }
    \label{tab:task3_detail}
\end{table}

\paragraph{Pick Bread}

This task involves picking up a slice of bread and placing it into a toaster, which requires control of the gripper to position the bread accurately without damaging it, as shown in Figure \ref{fig:tasks_detail}. The challenge lies in aligning the bread with the toaster's slot and lowering it, avoiding collisions with the toaster or the surrounding environment. We report more specific details of the policy training for this task in Table \ref{tab:task4_detail}. The task description for the VLA model is "Put the bread in the grey toaster."

\begin{table}[htbp]
    \centering
    \begin{tabular}{c|c}
        Parameter &Value\\
        \hline 
        Action space &7-dimensional\\
        Initial offline demonstrations &30\\
        Max episode length &100\\
        Reset method &Human reset\\
        Randomization range & 2 cm in x and y\\
        $(\alpha, \beta, \eta)$ for offline fine-tuning & $(0.01, 1.0, 0.1)$\\
        $(\beta, \eta)$ for online fine-tuning & $(0.5, 1.0)$\\
    \end{tabular}
    \caption{\textbf{Policy training details for the Pick Bread task.} }
    \label{tab:task4_detail}
\end{table}

\paragraph{Open Toaster}

This task involves pressing the button on a toaster to start the toasting process, as shown in Figure \ref{fig:tasks_detail}. It requires precise control of the gripper to avoid slipping or applying excessive force while ensuring that the button is pressed in a controlled and consistent manner. We report more specific details of the policy training for this task in Table \ref{tab:task5_detail}. The task description for the VLA model is "Press the black button and open the toaster."

\begin{table}[htbp]
    \centering
    \begin{tabular}{c|c}
        Parameter &Value\\
        \hline 
        Action space &6-dimensional\\
        Initial offline demonstrations &20\\
        Max episode length &100\\
        Reset method &Script reset\\
        Randomization range & 2 cm in y and z\\
        $(\alpha, \beta, \eta)$ for offline fine-tuning & $(0.01, 1.0, 0.1)$\\
        $(\beta, \eta)$ for online fine-tuning & $(0.5, 1.0)$\\
    \end{tabular}
    \caption{\textbf{Policy training details for the Open Toaster task.} }
    \label{tab:task5_detail}
\end{table}

\paragraph{Put Bread}

This task involves picking up a slice of toasted bread from the toaster and placing it on a white plate, as shown in Figure \ref{fig:tasks_detail}. The challenge lies in the precision required to grasp the toast without crushing or damaging it. The gripper must carefully move the toast from the toaster slot while avoiding contact with the toaster's edges or other objects nearby. We report more specific details of the policy training for this task in Table \ref{tab:task6_detail}. The task description for the VLA model is "Put the bread on the white plate."

\begin{table}[htbp]
    \centering
    \begin{tabular}{c|c}
        Parameter &Value\\
        \hline 
        Action space &7-dimensional\\
        Initial offline demonstrations &30\\
        Max episode length &120\\
        Reset method &Human reset\\
        Randomization range & 2 cm in x and y\\
        $(\alpha, \beta, \eta)$ for offline fine-tuning & $(0.01, 1.0, 0.1)$\\
        $(\beta, \eta)$ for online fine-tuning & $(0.5, 1.0)$\\
    \end{tabular}
    \caption{\textbf{Policy training details for the Put Bread task.} }
    \label{tab:task6_detail}
\end{table}

\paragraph{Insert Wheel}

This task involves installing wheels on the chair base by inserting pins into their corresponding slots, as shown in Figure \ref{fig:tasks_detail}. It is a contact-rich task requiring precise control to ensure the pins align correctly with the slots. The complexity of this task increases due to the tight tolerances and complex contact dynamics between the pin and the slot, making it a highly demanding task that requires precision and control. We report more specific details of the policy training for this task in Table \ref{tab:task8_detail}. The task description for the VLA model is "Insert the black wheel into the grey chair base."

\begin{table}[htbp]
    \centering
    \begin{tabular}{c|c}
        Parameter &Value\\
        \hline 
        Action space &7-dimensional\\
        Initial offline demonstrations &30\\
        Max episode length &100\\
        Reset method &Human reset\\
        Randomization range & 2 cm in x and y\\
        $(\alpha, \beta, \eta)$ for offline fine-tuning & $(0.01, 1.0, 0.1)$\\
        $(\beta, \eta)$ for online fine-tuning & $(0.5, 1.0)$\\
    \end{tabular}
    \caption{\textbf{Policy training details for the Insert Wheel task.} }
    \label{tab:task8_detail}
\end{table}

\paragraph{Hang Chinese Knot}

This task involves hanging a Chinese knot on a hook, which requires careful manipulation of a soft and dynamic object, as shown in Figure \ref{fig:tasks_detail}. The task requires fine dexterity to handle the knot's soft body and to maintain its structure while attaching it to the hook. The task involves dealing with the dynamics of soft object manipulation, where maintaining consistent contact and proper tension is critical for success. We report more specific details of the policy training for this task in Table \ref{tab:task9_detail}. The task description for the VLA model is "Hang the Chinese knot on the hook."

\begin{table}[ht]
    \centering
    \begin{tabular}{c|c}
        Parameter &Value\\
        \hline 
        Action space &7-dimensional\\
        Initial offline demonstrations &30\\
        Max episode length &100\\
        Reset method &Human reset\\
        Randomization range & 3 cm in y and z\\
        $(\alpha, \beta, \eta)$ for offline fine-tuning & $(0.01, 1.0, 0.1)$\\
        $(\beta, \eta)$ for online fine-tuning & $(0.5, 1.0)$\\
    \end{tabular}
    \caption{\textbf{Policy training details for the Hang Chinese Knot task.} }
    \label{tab:task9_detail}
\end{table}

\subsection{More experiment results}

\label{apx:more_exp}
\begin{figure*}[ht]
\centering
\includegraphics[width=\linewidth]{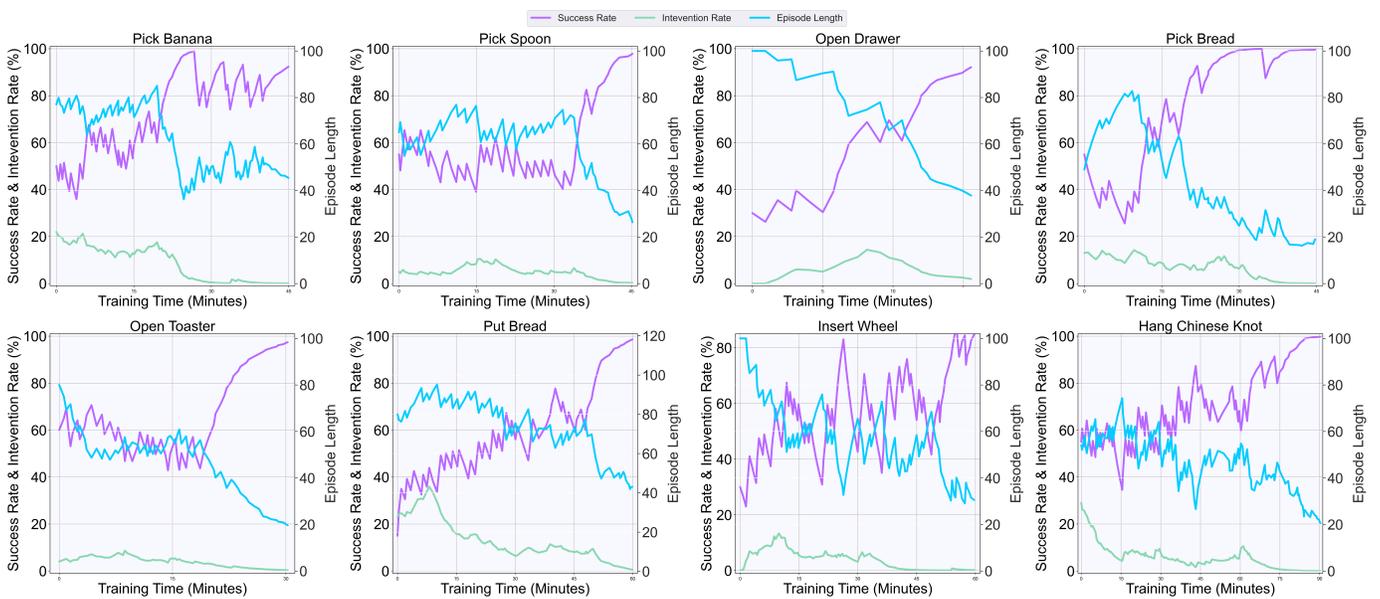}
\caption{\textbf{Learning curves during online training for all tasks.} This figure presents the success rates, intervention rates, and episode lengths, displayed as a running average of over 20 episodes.}
\label{fig:all_result}
\end{figure*}

In this section, we provide all policy learning curves for HIL-ConRFT on all tasks in Figure \ref{fig:all_result}. 

\end{document}